\documentclass[10pt,twocolumn,letterpaper]{article}

\usepackage[pagenumbers]{cvpr} 

\newif\ifshowcomments
\showcommentsfalse     

\showcommentstrue
\usepackage{colortbl}
\definecolor{tabfirst}{rgb}{1, 0.7, 0.7} 
\definecolor{tabsecond}{rgb}{1, 0.85, 0.7} 
\definecolor{tabthird}{rgb}{1, 1, 0.7} 

\usepackage{pifont}
\newcommand{\cmark}{\ding{51}} 
\newcommand{\xmark}{\ding{55}} 

\usepackage{multirow}
\usepackage{makecell}
\usepackage{enumitem}
\usepackage[percent]{overpic}
\usepackage{lipsum}
\usepackage{xcolor}
\usepackage{annotate-equations}
\usepackage{diagbox}
\usepackage{tikz}
\usetikzlibrary{decorations.pathreplacing}

\usepackage{tcolorbox}
\tcbuselibrary{skins, breakable}

\newtcolorbox[auto counter, number within=section]{PromptBox}[3][]{
  sharp corners,
  boxrule=0.6pt,
  fonttitle=\bfseries,
  colback=#2!5,       
  colframe=#2!50,     
  title={\thetcbcounter~~#3}, 
  left=6pt,
  right=6pt,
  top=6pt,
  bottom=6pt,
  before skip=10pt,
  after skip=10pt,
  parbox=false,
  #1
}

\newcommand{\methodname}{\textit{Over\texttt{++}}}

\newcommand{\inoeffect}{\ensuremath{\mathcal{I}_{\text{over}}}}
\newcommand{\meffect}{\ensuremath{\mathcal{M}_{\text{effect}}}}
\newcommand{\igt}{\ensuremath{\mathcal{I}_{\text{gt}}}}
\newcommand{\ifg}{\ensuremath{\mathcal{I}_{\text{fg}}}}
\newcommand{\istarfg}{\ensuremath{\mathcal{I}^*_{\text{fg}}}}
\newcommand{\ibg}{\ensuremath{\mathcal{I}_{\text{bg}}}}
\newcommand{\msubject}{\ensuremath{\mathcal{M}_\text{subject}}}

\let\svthefootnote\thefootnote
\newcommand\freefootnote[1]{%
\let\thefootnote\relax%
\footnotetext{\hskip -14pt #1}%
\let\thefootnote\svthefootnote%
}

\usepackage{tocloft} \usepackage{xcolor}  
 
\definecolor{cvprblue}{rgb}{0.21,0.49,0.74}
\usepackage[pagebackref,breaklinks,colorlinks,allcolors=cvprblue]{hyperref}

\def\confName{CVPR}
\def\confYear{2026}

\title{Over++: Generative Video Compositing for Layer Interaction Effects}

\author{
    Luchao Qi$^{1}$\textsuperscript{*}\hspace{0.5em}
    Jiaye Wu$^{2}$\hspace{0.5em}
    Jun Myeong Choi$^{1}$\hspace{0.5em}
    Cary Phillips$^{3}$\hspace{0.5em}
    Roni Sengupta$^{1}$\hspace{0.5em}
    Dan B Goldman$^{3}$\\[0.1em]   
    $^{1}$University of North Carolina at Chapel Hill 
    \quad $^{2}$University of Maryland 
    \quad $^{3}$Industrial Light \& Magic
    \\[0em]   
    {\tt\normalsize
    Project Page: \href{https://overplusplus.github.io/}{https://overplusplus.github.io}
    }
}

\begin{document}
\twocolumn[{%
\renewcommand\twocolumn[1][]{#1}%
\maketitle
\vspace{-1em}
\begin{overpic}[width=\linewidth]{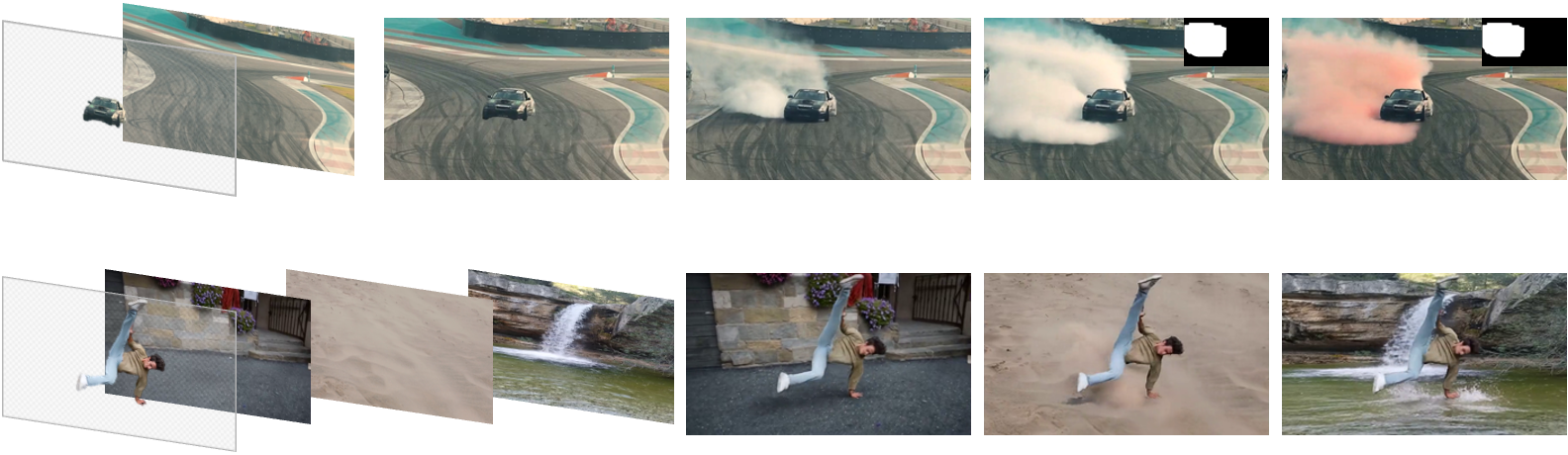}

    \def\ypos{29.4}
    \put(0, \ypos){\small Foreground}
    \put(12, \ypos){\small Background}
    \put(27.5, \ypos){\small Input w/o effects}
    \put(46.8, \ypos){\small Output w/ effects}
    \put(64.5, \ypos){\small Effect control: mask}
    \put(82.5, \ypos){\small Effect control: prompt }

    \put(7.5, 11.8){%
    \begin{tikzpicture}
      \draw [decorate,decoration={brace,amplitude=4pt}]
            (0,0) -- (6,0)
            node[midway,above,yshift=4pt]{};
    \end{tikzpicture}
    }
    
    \def\ypos{13}
    \put(0 , \ypos){\small Foreground}
    \put(16, \ypos){\small Multiple backgrounds (BG)}
    \put(45.7, \ypos){\small Output (over BG 1)}
    \put(64.5, \ypos){\small Output (over BG 2)}
    \put(83.7, \ypos){\small Output (over BG 3)}

\end{overpic}

\captionof{figure}{
\textbf{\methodname.}
Our video generation model synthesizes environmental effects between foreground and background layers of a video composite.
\textbf{Top:} \methodname~supports versatile effect control, such as guiding a smoke effect with a mask (2nd from right), and/or modifying it with a text prompt (\textit{``Red smoke''} at far right).
\textbf{Bottom:} Given multiple background options, \methodname~can generate diverse context-aware effects, such as shadows for BG~1, dust and shadows for BG~2, and splashes and reflections for BG~3.
} 
\vspace{1em}
\label{fig:teaser}
}]



\begin{abstract}
\freefootnote{\textsuperscript{*}Work done during an internship at Industrial Light \& Magic.}

In professional video compositing workflows, artists must manually create environmental interactions—such as shadows, reflections, dust, and splashes—between foreground subjects and background layers. 
Existing video generative models struggle to preserve the input video while adding such effects, and current video inpainting methods either require costly per-frame masks or yield implausible results.
We introduce \textbf{augmented compositing}, a new task that synthesizes realistic, semi-transparent environmental effects conditioned on text prompts and input video layers, while preserving the original scene.
To address this task, we present \methodname, a video effect generation framework that makes no assumptions about camera pose, scene stationarity, or depth supervision. We construct a paired effect dataset tailored for this task and introduce an unpaired augmentation strategy that preserves text-driven editability. Our method also supports optional mask control and keyframe guidance without requiring dense annotations. 
Despite training on limited data, \methodname~produces diverse and realistic environmental effects and outperforms existing baselines in both effect generation and scene preservation.

\end{abstract}

\section{Introduction}

 In 1984, Porter and Duff~\cite{porterduff1984} formalized the ``Algebra of Compositing," defining the ``over" operator to combine image elements with a pre-multiplied alpha channel. Today, a visual effects compositor may construct a vide clip using dozens or even hundreds of layers, starting from filmed or computer-generated background and foreground elements, and gradually inserting additional elements to marry these key elements together. Such additional elements may include physically-based environmental interactions such as shadows, reflections, dust, and water splashes (Fig.~\ref{fig:teaser}, bottom). 
 In this work, we envision the application of generative video for compositing. We call our approach \methodname, in homage to Porter and Duff's seminal paper.

\begin{figure}[t]
\vspace{1em}
\centering
\begin{overpic}[width=\linewidth]{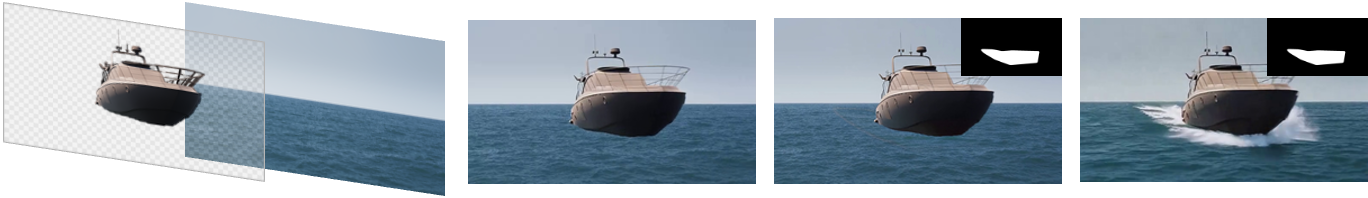}
    \def\ypos{15.5}
    \put(0,\ypos){\scriptsize Foreground}
    \put(16,\ypos){\scriptsize Background}
    \put(35,\ypos){\scriptsize Input w/o effects}
    \put(60,\ypos){\scriptsize VACE~\cite{vace}}
    \put(79,\ypos){\scriptsize Output w/ effects}
\end{overpic}


\captionof{figure}{
\textbf{Limitations of inpainting models for effects.}
Simply compositing foreground ``over'' background produces an input without effects.  
Inpainting models such as VACE~\cite{vace} require per-frame mask and may still fail to generate the desired effect.
Our method successfully produces the target wake (far right).
}
\label{fig:subteaser}

\vspace{-1em}
\end{figure}

Traditional video compositing tools such as Foundry’s Nuke~\cite{nuke} and Adobe After Effects~\cite{adobeae} offer nondestructive editing pipelines across sequences of shots, but still require substantial manual effort from artists.
Modern generative video models can produce highly realistic content, yet their stochastic behavior prevents them from integrating into the controlled, iterative workflows that professional compositors rely on.
Recent inpainting-based approaches (e.g.,~\cite{vace}) allow users to specify regions for modification via masks, but, as shown in Fig.~\ref{fig:subteaser}, they struggle to generate complex environmental effects such as wakes and require labor-intensive per-frame mask annotations.

 In this work, we propose \textit{augmented compositing}: a user provides a foreground and background element, a text prompt describing an environmental interaction, and an optional mask video specifying the regions in which the interaction should appear. The system automatically produces a video adhering to the user's desired effects. This can be seen as a variation of inpainting that is specialized for semi-transparent or stochastic effects which do not replace or fundamentally alter the other elements. This behavior is critical for professional use cases for two reasons: First, because it makes it possible to ensure continuity of both the subject's and environment's appearance across a sequence. Second, it adheres more accurately to the directorial intent of the filmmakers who acquired the input footage.
 


We address the task by introducing a data collection pipeline that generates paired synthetic videos (with and without effects), paired real-world videos (with and without effects), and unpaired videos (with effects). 
Our approach extends prior video generative models with a training strategy that produces effects under optional mask guidance while preserving core instruction-following capabilities. 
Despite the limited training set (54 real-world paired, 573 synthetic paired, and 460 unpaired videos), \methodname~generalizes effectively to diverse environmental effects, including shadows, dust, smoke, and water splashes (Fig.~\ref{fig:teaser}).
Our contributions include: (i) the introduction of \emph{augmented compositing} as a new generative task, (ii) \methodname, a video effect generation model fine-tuned on the newly constructed dataset, and (iii) a comprehensive evaluation with quantitative and qualitative comparisons to prior work, including metrics tailored for this task.


\section{Related Work}
\label{sec:related_work}
\noindent \textbf{VFX Generation.}
Visual effects (VFX) generation plays a crucial role in modern video production. Traditional VFX creation often relies on physics-based simulations in digital content creation tools such as \emph{Houdini} or \emph{Maya}, or on manual compositing of pre-existing effect elements in software such as \emph{Nuke} or \emph{After Effects}~\cite{nuke,adobeae}. These approaches ensure physical plausibility and visual realism but are computationally expensive, time-consuming, and require substantial manual effort and artistic expertise. We refer the reader to the VES Handbook~\cite{okun2010ves} for additional details.
Recent works~\cite{liu2025vfx, mao2025omni, 10.1145/3664647.3681516} have explored controllable VFX generation using generative models, either without effect references~\cite{mao2025omni} or with reference exemplars~\cite{10.1145/3664647.3681516}. These efforts aim to simplify VFX creation by leveraging generative AI to replace or complement manual simulation. However, they primarily target stylized or exaggerated effects~\cite{li_vfxmaster_2025}, and do not explicitly model physically grounded environmental interactions between objects and their surroundings.

To bridge the gap between realism and efficiency, recent works~\cite{chen_physgen3d_2025, liu2024physgen, tan2024physmotion} have explored integrating physical simulation with generative modeling, enabling controllable and physically plausible interactions. 
For instance, PISA~\cite{li2025pisa} leverages simulated videos to fine tune generative models, primarily focusing on object-level phenomena such as free fall. 
While these approaches demonstrate promise, they are typically constrained to simple, isolated object dynamics. 
Extending simulation-driven generation to scenes with multiple interacting objects of different types is computationally expensive, and it remains costly and difficult to make perceptually convincing simulations of non-rigid or volumetric effects such as dust, fluids, and smoke.

\noindent \textbf{Omnimatte.} 
Omnimatte~\cite{lu_omnimatte_2021} and related methods aim to decompose a visual input into RGBA matte layers, each containing an object and its associated effects.  
In the image domain, recent works~\cite{zhao_objectclear_2025, yang2025generative, chen2024freecompose, li2024RORem} decompose an image into a clean background layer and a transparent foreground layer that preserves secondary visual effects such as shadows and reflections.  
Magic Fixup~\cite{alzayer_magic_2025} further enables object-level editing while adaptively adjusting the associated effects.  
In the video domain, follow-up methods~\cite{lee_generative_2024, miao_rose_2025} extend this decomposition to video layers, while OmnimatteZero~\cite{samuel_omnimattezero_2025} advances toward real-time performance.

Although such decomposition-based approaches can be regarded as the inverse of our problem—separating rather than composing layers—they inherently lack explicit controllability. In contrast, our goal is to compose foreground and background videos under the joint guidance of a spatial mask and a textual prompt, enabling controllable synthesis of physically grounded environmental effects.

\noindent \textbf{Visual Concept Composition.}
Visual concept composition aims to merge multiple reference inputs into a coherent output. 
Recent image- and video-based approaches~\cite{chen2025humo,wang2025dreamactor,sang2025lynx,liu2025phantom,huang2025conceptmaster,chen2025contextflow} leverage advances in generative modeling to combine visual content and style from reference inputs. 
Beyond static inputs, dynamic concept methods~\cite{abdal2025dynamic,abdal2025zero,yang2025gencompositor} compose content across videos, either by fine-tuning personalized models (e.g., DreamBooth~\cite{ruiz_dreambooth_2023_fixed}, Textual Inversion~\cite{gal_image_2022}) or by parameter-efficient adaptation using LoRA~\cite{hu2022lora}. 

While these methods achieve impressive visual coherence, our work targets a distinct problem setting: given separate foreground and background layers, we synthesize \emph{environmental effects}—such as shadows, splashes, or smoke—that perceptually connect the two layers without altering their original content or motion. 
This formulation differs from prior composition methods, which primarily focus on blending appearance or identity, often at the cost of foreground or background fidelity—a key requirement in real-world compositing workflows.

\noindent \textbf{Video Generation and Control.}
Recent advances in video diffusion models have substantially improved video generation quality and controllability~\cite{gao_seedance_2025, bar2024lumiere, xiong2025talkingheadbench}. Existing approaches provide various forms of conditioning, including text-to-video (T2V)~\cite{wan_wan_2025, yang_cogvideox_2024, HaCohen2024LTXVideo} and image-to-video (I2V)~\cite{wang2025dreamvideo, ren2024consisti2v, singer2022make}.
Force-Prompting~\cite{gillman_force_2025} extends this line by fine-tuning an I2V backbone with force-based conditioning, enabling simulation of object–environment interactions such as wind acting on fabric.
Recent works further explore joint image–video conditioning (I+V2V)~\cite{zhao2024motiondirector}, where I2VEdit~\cite{ouyang2024i2vedit} propagates first-frame edits across the entire sequence for temporally consistent control.
Unified models such as UNIC~\cite{ye2025unic} and VACE~\cite{vace} aim to generalize across multiple video editing tasks—including insertion, deletion, and camera control—within a single framework.

While these models achieve impressive controllability and visual fidelity, they primarily target object- or appearance-level editing rather than the generation of environmental effects. ActAnywhere~\cite{NEURIPS2024_34a9582c_fixed} synthesizes plausible effects given an edited background image; however, it is not directly applicable to professional compositing scenarios that require preserving a dynamic background video, which our method explicitly supports.





\begin{figure*}[th]
    \centering
    \begin{overpic}[width=1\linewidth]{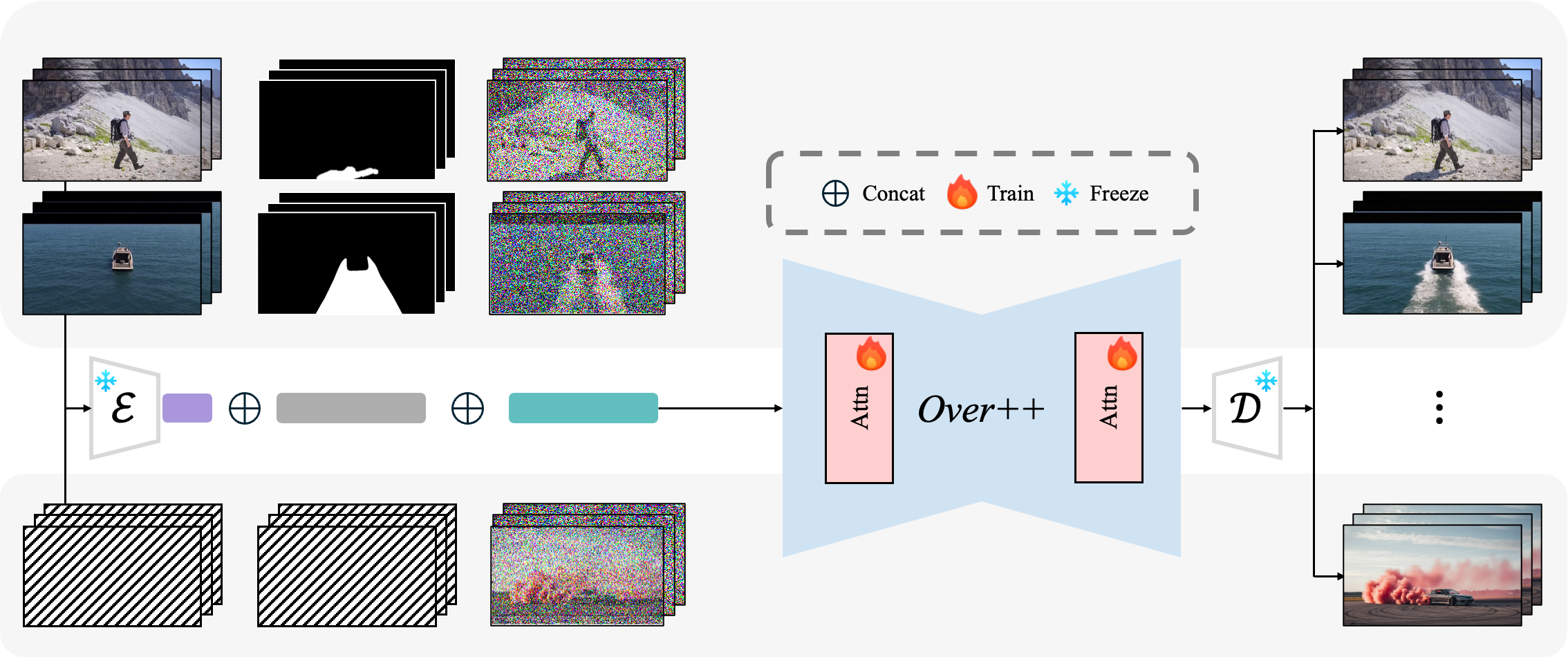}

    \def\ypos{39.5}
    \put(1.5,\ypos){\scriptsize Input w/o effects $\inoeffect$}
    \put(17,\ypos){\scriptsize Effect mask $\meffect$}
    \put(34,\ypos){\scriptsize Noisy video}
    \put(86,\ypos){\scriptsize Output w/ effects $\hat{\mathcal{I}}$}

    \put(50,36){\normalsize Paired video training (Sec.~\ref{sec:training_data})}
    \put(49, 2){\normalsize Unpaired video training (Sec.~\ref{sec:control_prompt})}

    \put(64,13){\scriptsize Eq.~\ref{eq:method}}
    
    \end{overpic}
    
    \caption{
    \textbf{\methodname~framework.}
    Given an input composite video lacking environmental effects such as shadows or wakes ($\inoeffect$), and an optional binary mask indicating the target effect regions ($\meffect$), our model \methodname~generates desired effects within the specified regions ($\hat{\mathcal{I}}$). 
    Training includes unpaired data by zeroing out the latent codes of $\inoeffect$ and $\meffect$. (Text prompts $\mathcal{T}$ are not shown here for simplicity.)
    }
    \label{fig:method}

\end{figure*}

\section{Method}

In this section, we formulate augmented compositing as a variation of the video inpainting problem (Sec.~\ref{sec:task_formulation}) and introduce our method, \methodname. We treat effect synthesis primarily as a supervised learning task and construct paired training videos with and without effects (Sec.~\ref{sec:training_data}). To enable fine-grained control over generated effects, we further incorporate spatial masking and text prompts (Sec.~\ref{sec:control_effect}). Although trained mainly on synthetic data with limited real-world examples, \methodname~generalizes effectively to in-the-wild videos, producing realistic, prompt-guided effects within arbitrary human-annotated masks.

\subsection{Task Formulation}
\label{sec:task_formulation}
Video visual effects compositing aims to merge separate video elements—typically a foreground subject and a background environment—into a single coherent scene while preserving their original appearances. However, naively overlaying these elements using a simple ``over'' operation~\cite{porterduff1984} fails to reproduce essential environmental interactions between the subject and its surroundings, such as shadows, reflections, and splashes. Achieving visual realism therefore requires synthesizing these effects in spatially and temporally consistent regions of the composed scene.

To address this challenge, we extend the classic ``over'' formulation by introducing a masking mechanism that specifies where foreground–background interaction effects should appear, enabling fine-grained spatial control over effect placement. Given a simple RGB composite video ``foreground over background,'' denoted $\inoeffect$, an effect mask video $\meffect$ delineating the regions designated for effect generation, and a text prompt $\mathcal{T}$ describing the desired effect, our goal is to synthesize a video $\hat{\mathcal{I}}$ that preserves the original appearance and motion of $\inoeffect$ while adding new, visually plausible effects consistent with $\mathcal{T}$ within the masked regions of $\meffect$.
To this end, we train a video inpainting diffusion model $\mathcal{G}$, referred to as \methodname, which generates physically grounded effects conditioned jointly on $\meffect$ and $\mathcal{T}$.


\methodname~fine-tunes a base video inpainting diffusion transformer (DiT) model. During fine-tuning, we update all transformer attention blocks while keeping the VAE encoder and decoder frozen.
Unlike prior inpainting methods, which zero out masked regions in $\inoeffect$~\cite{li2025diffueraser,Xie2022SmartBrushTAA}, we pass through the fully-encoded latents to preserve scene context and suppress hallucinations. 
As shown in Fig.~\ref{fig:method}, \methodname~denoises Gaussian noise $\mathcal{N}$ to reconstruct the target video $\igt$, conditioned on the concatenated latents of the effect mask $\meffect$, the composite input video $\inoeffect$, and the textual description $\mathcal{T}$ via attention~\cite{vaswani_attention_2023}. Formally,
\begin{align}
    \igt \approx \hat{\mathcal{I}} = \mathcal{G}(\mathcal{N}; \inoeffect, \meffect, 
    \mathcal{T}),
    \label{eq:method}
\end{align}
Note that while professional compositing also involves complementary tasks such as motion alignment and color harmonization, these aspects are orthogonal to our scope and have been extensively studied in prior work on motion transfer~\cite{gu_diffusion_2025} and video harmonization~\cite{Harmonizer}.
In contrast, our work focuses exclusively on effect generation, a problem that remains comparatively underexplored.

\subsection{Dataset}
\label{sec:training_data}

\noindent
The core challenge of training \methodname~is the scarcity of paired videos $(\inoeffect, \igt)$ with and without effects. To address this, we leverage the Omnimatte methods~\cite{lee2025generative, lin2023omnimatterf, lu_omnimatte_2021} to decompose each video $\igt$ into separable layers. Given a video with effects $\igt$, Omnimatte methods decompose it into a foreground layer $\istarfg$ and a clean background layer $\ibg$, following the formulation 
$
\igt \approx \alpha \cdot \istarfg + (1 - \alpha) \cdot \ibg,
$
where $\alpha$ denotes the per-pixel alpha matte, and $\istarfg$ represents the video of the foreground subject $\ifg$ and its associated effects.
An example of such decomposition is shown in Fig.~\ref{fig:method_dataset} (bottom). Given a biker–puddle video ($\igt$), it can be decomposed into a foreground ($\istarfg$) containing the biker, splashes, and reflections, and a background ($\ibg$) showing the undisturbed puddle as if no interaction occurs.

We further extract the pure foreground subject $\ifg$ from the decomposed foreground $\istarfg$ using a subject mask $\msubject$ obtained from off-the-shelf segmentation tools~\cite{liu2024grounding,ravi2024sam2}. The extracted subject $\ifg$ is then re-composited over the clean background $\ibg$ using the standard `\emph{over}' operation~\cite{porterduff1984}, discarding the attached environmental effects. Formally,
\begin{align}
\inoeffect = \msubject \cdot \istarfg + (1 - \msubject) \cdot \ibg,
\label{eq:data_over}
\end{align}
We construct training videos from the following sources:  
(a) \textbf{54 real-world paired videos} created from DAVIS~\cite{pont20172017}, Pexels~\footnote{\href{https://www.pexels.com/}{https://www.pexels.com/}}, and prior works~\cite{lee2025generative, lin2023omnimatterf, samuel_omnimattezero_2025, lu_omnimatte_2021}, featuring complex effects such as smoke, water splashes, and reflections; 
(b) \textbf{573 synthetic paired videos} sourced from the Movies dataset~\cite{lin2023omnimatterf} and additional synthetic data rendered using Blender~\cite{blender} and Kubric~\cite{Greff_2022_CVPR}, following the procedures in~\cite{lee2025generative, lin2023omnimatterf}. These synthetic videos complement the limited real-world samples and introduce greater diversity in shadows and reflections, which are easier to simulate; and  
(c) \textbf{460 unpaired videos} generated from a pretrained T2V model. In addition to the paired videos from (a) and (b), we generate unpaired videos based on text prompts to produce diverse environmental effects. These unpaired T2V videos contain only $\igt$ and do not have corresponding $\meffect$ or $\inoeffect$. Training with such unpaired data helps preserve the pretrained model’s text-to-video (T2V) generation capabilities and further enhances the classifier-free guidance (CFG)–based prompt editing capacity of the final model. Details of T2V unpaired data (c) are discussed in Sec.~\ref{sec:control_prompt}.

\begin{figure}[t]
    \centering
    \begin{overpic}[width=\linewidth]{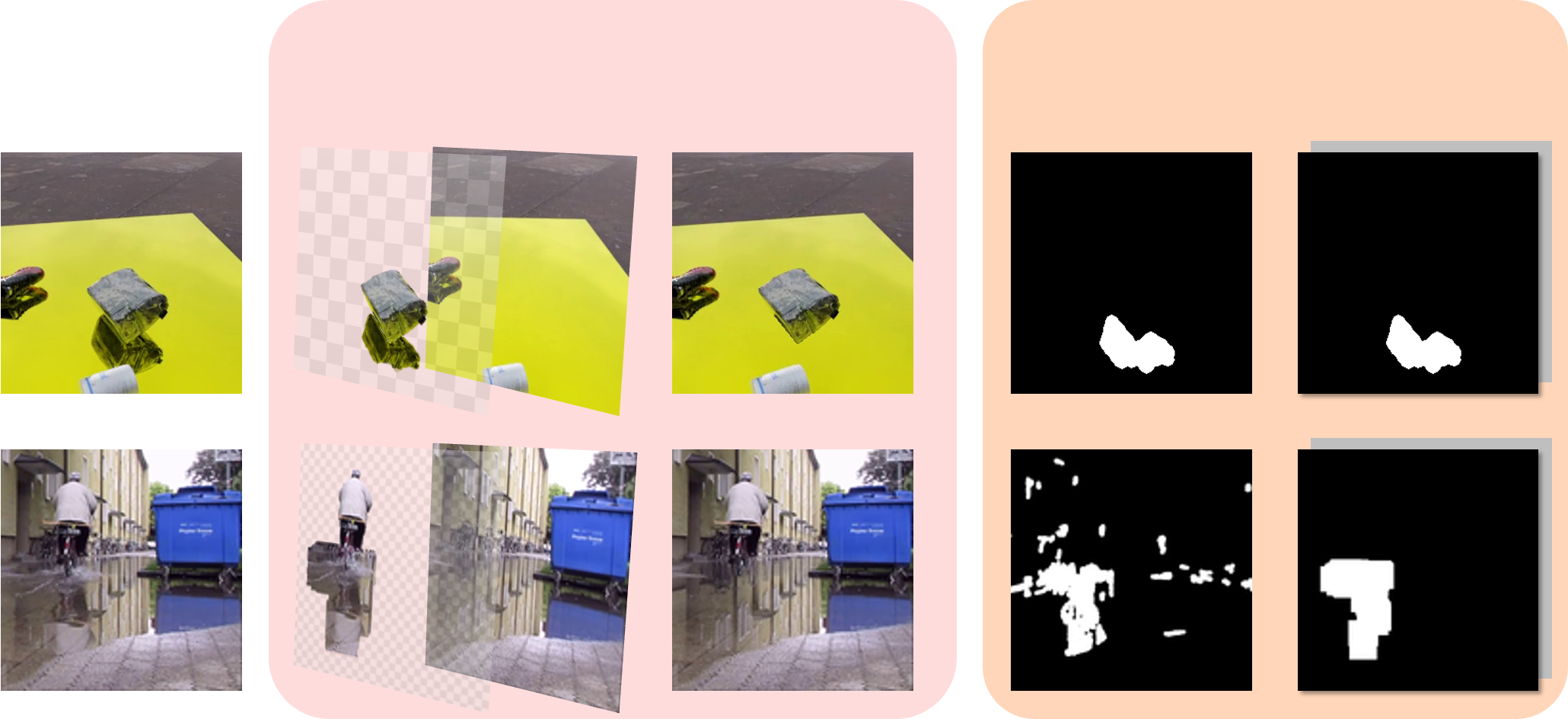}
    \def\ypos{42.5}
    \put(33.5, \ypos){\footnotesize Sec.~\ref{sec:training_data}}
    \put(75, \ypos){\footnotesize Sec.~\ref{sec:control_mask}}

    \def\ypos{38.5}
    \put(5,\ypos){\scriptsize $\igt$}
    \put(22,\ypos){\scriptsize $\istarfg$}
    \put(32,\ypos){\scriptsize $\ibg$}
    \put(43.2,\ypos){\scriptsize $\inoeffect$~ (Eq.~\ref{eq:data_over})}
    \put(64.7,\ypos){\scriptsize $\delta (\igt, \inoeffect)$}
    \put(86,\ypos){\scriptsize $\meffect$}

    \end{overpic}
    \caption{
    \textbf{Mask generation for training data.}
    Given a training video with effects $\igt$, we construct a version without effects $\inoeffect$.
    The effect mask $\meffect$ is derived by applying mask pruning to the difference image $\delta(\igt, \inoeffect)$ to remove noise and artifacts.
    \textbf{Top:} Synthetic data with clean $\delta (\igt, \inoeffect)$.
    \textbf{Bottom:} Real-world data has noisier $\delta (\igt, \inoeffect)$, requiring additional cleanup.
    }
    \label{fig:method_dataset}

\vspace{-1em}
\end{figure}


\subsection{Controlling Effect Generation}
\label{sec:control_effect}
\subsubsection{Mask-based Effect Generation}
\label{sec:control_mask}
VFX compositors often aim to generate effects within specific spatio-temporal regions, which we model using masks. 
Given the paired videos $(\inoeffect, \igt)$, we construct an effect mask $\meffect$ that localizes regions of effect occurrence in $\igt$ by computing the pixel-wise difference between the video pair, denoted as $\delta(\igt, \inoeffect)$. However, due to imperfections in the subject segmentation mask $\msubject$, VAE reconstructions, and video decomposition~\cite{zhao_objectclear_2025, gao_seedance_2025, seawead2025seaweed}, the decomposed foreground and background layers $(\istarfg, \ibg)$ may not be perfectly aligned with $\igt$. 
Such misalignment introduces pixel-level noise and minor artifacts in the computed difference, resulting in imperfect effect (Fig.~\ref{fig:method_dataset}, bottom).

To mitigate these artifacts, we first compute the pixel-wise difference between the collected pair, $\delta(\igt, \inoeffect)$, convert it to grayscale, and binarize it using Otsu’s thresholding~\cite{4310076}. 
The resulting binary mask is then refined through a sequence of morphological operations---erosion, dilation, and median filtering---to suppress salt-and-pepper noise and enhance spatiotemporal consistency. 
As shown in Fig.~\ref{fig:method_dataset}, this mask pruning process effectively removes noise from the initial effect mask, especially for real-world data. Despite these refinements, the resulting masks $\meffect$ may still be imperfect in precisely delineating effect regions at the pixel level. However, such imperfections act as a form of natural data augmentation, enhancing robustness to loosely drawn user masks. Consequently, \methodname~can operate effectively even when provided with coarse, hand-drawn masks---a property further demonstrated in \textit{Robust Mask Editing} (Sec.~\ref{sec:application}).


In real-world applications, the effect mask $\meffect$ often requires frame-by-frame annotation by professional VFX artists, which is impractical for many use cases. To overcome this limitation, we introduce a \textit{tri-mask} design that supports training under both masked and unmasked conditions. Specifically, during training, we augment $\meffect$ by randomly replacing it with a uniform gray region to represent frames where effect regions are unknown or unannotated, indicating the frame may or may not contain effects, or that the locations of effects are uncertain.
This design offers the following two key benefits: (i) it enables \methodname~to operate seamlessly in both supervised (mask-guided) and unsupervised (mask-free) settings, providing a unified framework for diverse user scenarios; (ii) it allows \methodname~to perform inference with mixed guidance within a single model—allowing certain keyframes to be annotated while other remains unannotated, as further demonstrated in \textit{Keyframe Annotation} (Sec.~\ref{sec:application}).


\subsubsection{Prompt-based Effect Generation}
\label{sec:control_prompt}


Users can generate diverse effects with text prompting. For instance, users may request \textit{turbulent} versus \textit{calm} splashes, \textit{soft} versus \textit{harsh} shadows, or \textit{red} versus \textit{white} smoke. Achieving such diversity demands strong prompt-editing capabilities from the model. 
However, when trained solely on the limited paired data described in Sec.~\ref{sec:training_data}, the model can suffer from language drift~\cite{pmlr-v119-lu20c, ruiz_dreambooth_2023_fixed}, losing its inherent text-to-video (T2V) generation and prompt-editing abilities after fine-tuning. Ideally, this could be mitigated by training on \textit{multiple} target videos $\igt$ exhibiting diverse effects for the \textit{same} input composite $\inoeffect$, but such data are challenging to obtain or synthesize in practice. 
To address this limitation, we preserve the pre-trained model’s intrinsic T2V generation capabilities using additional unpaired data during training, and leverage classifier-free guidance (CFG)~\cite{ho_classifier-free_2022} at inference, enabling both mask- and prompt-conditioned effect generation.

To provide this additional unpaired data, we augment the training captions $\mathcal{T}$ using a LLM to generate semantically diverse descriptions of the same scene. For each caption, the language model produces multiple variants describing alternative physical effects or visual attributes, while maintaining the underlying scene semantics. 
The system prompt used for this augmentation is provided in the supplementary material (SM).
We then use these augmented prompts to synthesize additional videos $\igt$ with a pre-trained T2V model, expanding the training corpus beyond the paired data. 
When training with these unpaired videos, we use only $\igt$ and the caption $\mathcal{T}$, zeroing out the latents of the missing mask $\meffect$ and input video $\inoeffect$ to preserve \methodname’s text-to-video generation capabilities within the unified training framework.




\subsection{Implementation Details}

All experiments shown here were fine-tuned from the video inpainting variant\footnote{\href{https://github.com/aigc-apps/VideoX-Fun}{https://github.com/aigc-apps/VideoX-Fun}} of the text-to-video (T2V) diffusion transformer model, CogVideoX-5B~\cite{yang_cogvideox_2024}. 

We obtain video captions $\mathcal{T}$ using MiniCPM-V-2.6~\cite{yao2024minicpm} to generate dense spatio-temporal captions, which are then refined with LLaMA-3.1-8B-Instruct~\cite{grattafiori2024llama} into concise video-level descriptions (system prompt in the supplementary material).  
Unpaired training videos are generated with CogVideoX-5B from GPT-5–augmented prompts (Sec.~\ref{sec:control_prompt}).  
The rendered dataset and video processing code will be released upon acceptance.

We train \methodname~with a standard L2 diffusion loss~\cite{ho_denoising_2020}, on 384$\times$672 resolution (the output resolution of gen-omnimatte~\cite{lee_generative_2024}) using 8 NVIDIA A6000 GPUs for one day, totaling 1{,}000 iterations.
During inference, we apply temporal multidiffusion~\cite{Zhang_2024_CVPR} for videos longer than 85 frames, enabling effect generation on extended sequences.


\begin{figure*}[th]
\noindent
\begin{minipage}{\textwidth}

\centering
\captionof{table}{
\textbf{Quantitative comparison.} 
We evaluate effect generation performance on 24 videos at both the image and video levels. 
*~indicates methods that require an edited first frame for reference, where we use the first frame of the ground-truth video~$\igt$ with the added effects. 
Methods marked in {\setlength{\fboxsep}{1pt}\colorbox{gray!10}{\strut gray}} require masks. 
Best results are highlighted in  {\setlength{\fboxsep}{1pt}\colorbox{tabfirst}{\strut red}}, and second-best in {\setlength{\fboxsep}{1pt}\colorbox{tabsecond}{\strut orange}}. 
Please see SM for video results.
}
\resizebox{\textwidth}{!}{%
\begin{tabular}{lccccccccc}
    \toprule
    \multirow{2.5}{*}{Method} &
    \multicolumn{3}{c}{Image Eval} &
    \multicolumn{3}{c}{Image Eval} &
    \multicolumn{3}{c}{Video Eval} \\
    \cmidrule(lr){2-4} \cmidrule(lr){5-7} \cmidrule(lr){8-10}
    & $\text{CLIP}_{dir}~\uparrow$ & $\text{CLIP}_{text}~\uparrow$ & $\text{CLIP}_{img}~\uparrow$ &
    SSIM~$\uparrow$ & PSNR~$\uparrow$ & LPIPS~$\downarrow$ &
    FVD~$\downarrow$ & VMAF~$\uparrow$ & VBench~$\uparrow$ \\
    \midrule
    AnyV2V~\cite{ku2024anyv2v}~* & 21.05 & 31.22  & 84.31 & 0.57 & 17.38 & 0.27 & 786 & 12.01 & 0.181\\
    LoRA-Edit~\cite{gao_lora-edit_2025}~* & 37.72 & 30.72 & 88.49 & 0.39 & 12.89 & 0.39 & 970 & 6.65 & 0.181 \\
    Runway Aleph & 33.87 & 30.84 & 90.55 & 0.53 & 16.61 & 0.31 & 1297 & 5.44 & 0.181 \\
    Ours$^\dagger$ (w/o $\meffect$) & \cellcolor{tabsecond}43.49 & \cellcolor{tabsecond}31.56 & \cellcolor{tabsecond}95.25 & \cellcolor{tabfirst}0.80 & \cellcolor{tabsecond}23.58 & \cellcolor{tabfirst}0.13 & 608 & \cellcolor{tabsecond}28.19 & \cellcolor{tabfirst}0.188\\
    \midrule
    \cellcolor{gray!10}
    VACE-WAN2.1~\cite{vace} & 25.06 & 31.36 & 94.19 & 0.69 & 19.90 & 0.16 & \cellcolor{tabfirst}605 & 19.56 & 0.186 \\
    \cellcolor{gray!10}
    Ours w/o unpaired data & 42.93 & 29.89 & 95.34 & \cellcolor{tabfirst}0.80 & 22.89 & \cellcolor{tabfirst}0.13 & 612 & 28.05 & \cellcolor{tabfirst}0.188\\
    \cellcolor{gray!10}
    Ours (\methodname) & \cellcolor{tabfirst}46.27 & \cellcolor{tabfirst}31.58 & \cellcolor{tabfirst}95.48  & \cellcolor{tabfirst}0.80 & \cellcolor{tabfirst}23.75 & \cellcolor{tabfirst}0.13 & \cellcolor{tabfirst}605 & \cellcolor{tabfirst}29.30 & \cellcolor{tabfirst}0.188 \\
    \bottomrule
\end{tabular}%
}
\label{tab:benchmark}

\centering

\vspace{2em}
\begin{overpic}[width=\linewidth]{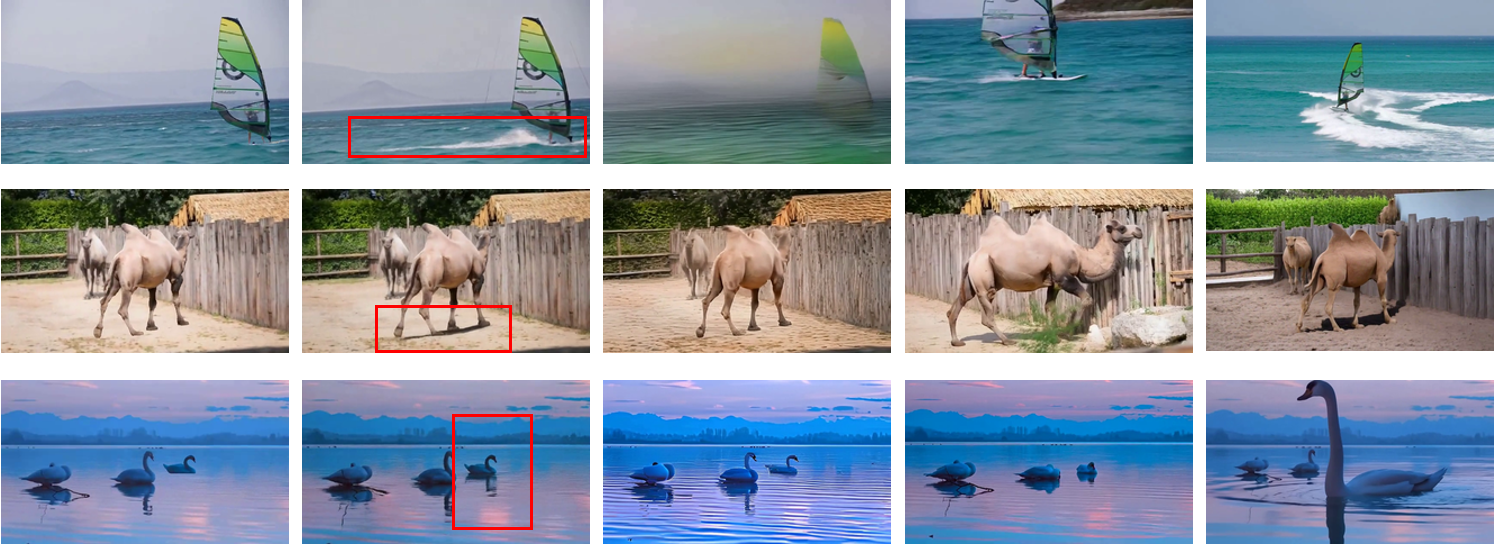}
\def\ypos{37}
\put(1,\ypos){\small Input w/o effects ($\inoeffect$)}
\put(21.7,\ypos){\small Ours$^\dagger$ (w/o $\meffect$)}
\put(45,\ypos){\small AnyV2V~\cite{ku2024anyv2v}}
\put(64.2,\ypos){\small LoRA-Edit~\cite{gao_lora-edit_2025}}
\put(85,\ypos){\small Runway Aleph}
\end{overpic}
\captionof{figure}{
\textbf{Visual comparison of effect generation without mask guidance.} 
We compare our model with state-of-the-art methods~\cite{ku2024anyv2v, gao_lora-edit_2025} and the commercial software Runway~Aleph. Generated effects are highlighted in {\setlength{\fboxrule}{1pt}\setlength{\fboxsep}{0pt}\fcolorbox{red}{white}{\strut red}} for our results. 
Note that alternate methods significantly change the identity, appearance, and sometimes visual composition of the source. Ours$^\dagger$ successfully generates the desired effects (top to bottom: wake, shadow, and reflection) while preserving the original input video subjects and composition.
}
\label{fig:benchmark}

\vspace{-1em}
\end{minipage}
\end{figure*}

\section{Results}

\begin{figure*}[th]
\centering

\begin{overpic}[width=\linewidth]{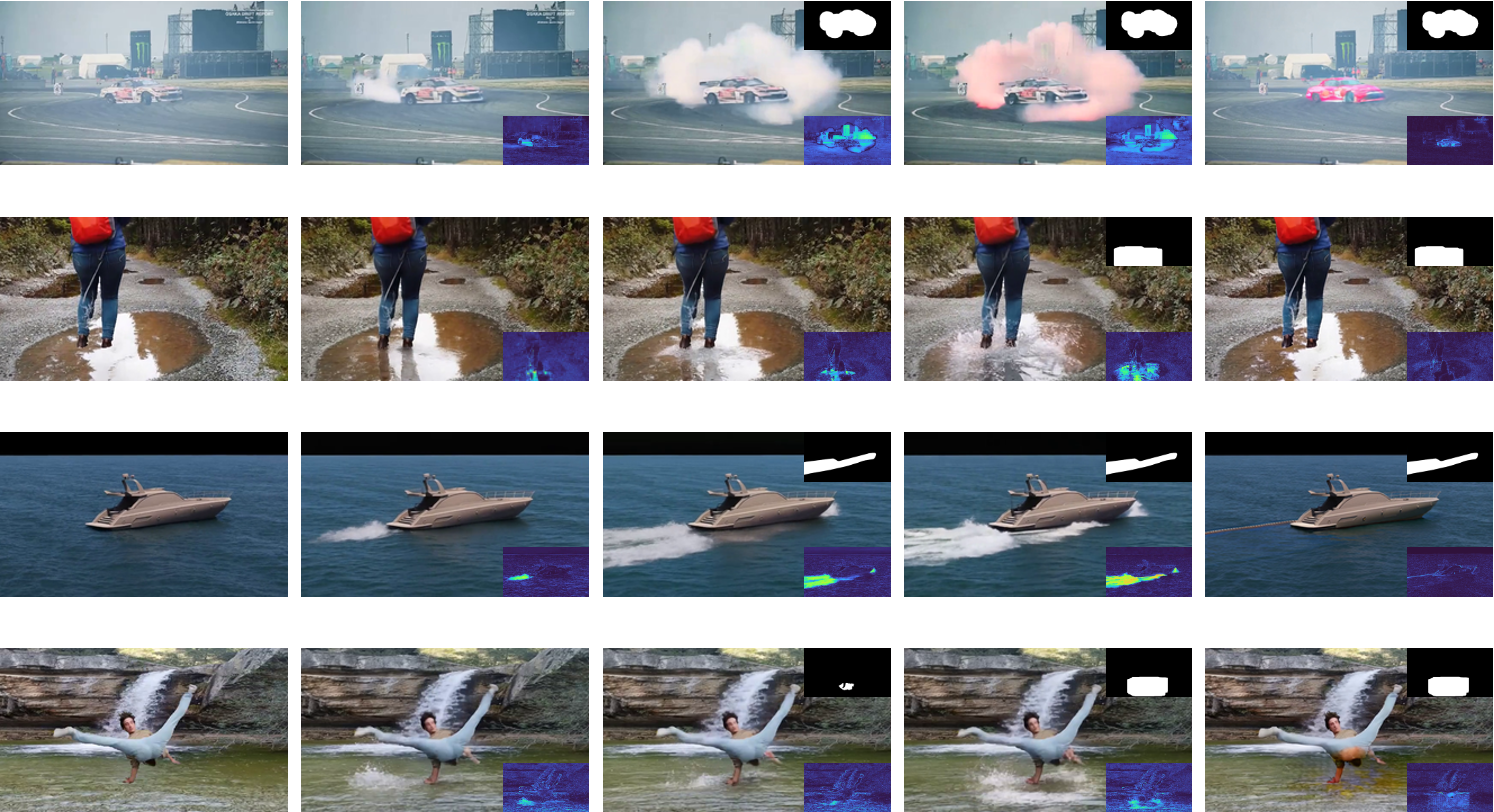}

\def\xposi{4}
\def\xposii{24.5}
\def\xposiii{45}
\def\xposv{86.5}

\def\ypos{55.1}
\put(\xposi,\ypos){\small Input w/o effects}
\put(\xposii,\ypos){\small Ours w/o mask}
\put(\xposiii,\ypos){\small Ours w/ mask}
\put(60,\ypos){\small Ours w/ prompt \textit{`Red smoke'}}
\put(\xposv,\ypos){\small VACE~\cite{vace}}

\def\ypos{40.7}
\put(\xposi,\ypos){\small Input w/o effects}
\put(\xposii,\ypos){\small Ours w/o mask}
\put(40, \ypos){\small Ours w/ CFG `More splash'}
\put(65,\ypos){\small Ours w/ mask}
\put(\xposv,\ypos){\small VACE~\cite{vace}}

\def\ypos{26.3}
\put(\xposi,\ypos){\small Input w/o effects}
\put(\xposii,\ypos){\small Ours w/o mask}
\put(\xposiii,\ypos){\small Ours w/ mask}
\put(60.3 ,\ypos){\small Ours w/ CFG \small `More wake'}
\put(\xposv,\ypos){\small VACE~\cite{vace}}

\def\ypos{11.7}
\put(\xposi,\ypos){\small Input w/o effects}
\put(\xposii,\ypos){\small Ours w/o mask}
\put(42,\ypos){\small Ours w/ mask (small)}
\put(62.5,\ypos){\small Ours w/ mask (large)}
\put(\xposv,\ypos){\small VACE~\cite{vace}}
\end{overpic}
\captionof{figure}{
\textbf{Extensions and downstream applications.} 
Each row shows a workflow enabled by \methodname~for effect generation on input videos without effects. 
\methodname~supports controllable generation under \emph{mask guidance}, \emph{text prompt guidance}, and \emph{CFG scaling}, applied in any order. 
For visualization, the mask $\meffect$ is shown at the top-right of each output, and the difference map $\delta(\inoeffect, \text{Output})$ at the bottom-right. 
We also compare against VACE~\cite{vace} using the same mask $\meffect$. 
Test videos are drawn (top to bottom) from DAVIS~\cite{pont20172017}, in-the-wild Pexels clips, synthetic CG data, and naive composites combining $\ifg$ and $\ibg$ from different sources. 
\methodname~successfully generates the desired effects (top to bottom: smoke, splash \& reflection, wake, and splash) with consistent quality and controllable variation.
}
\label{fig:control}

\vspace{-1em}
\end{figure*}

\subsection{Qualitative Comparison}
To the best of our knowledge, no prior method explores the specific problem setting we address. Therefore, we compare against the most closely related approaches and adapt them where necessary. 
Nonetheless, we emphasize that the core contribution of this paper lies in our new problem formulation, unified pipeline design, and training data engineering.

We compare our method against the following baselines:  
(a) \textbf{AnyV2V}~\cite{ku2024anyv2v} — a tuning-free image+video-to-video (I+V2V) framework that performs per-frame DDIM~\cite{song_denoising_2022} inversion and edits videos based on the inverted features and an edited first frame;  
(b) \textbf{VACE}~\cite{vace} — an all-in-one framework for video creation and editing. We evaluate its masked video-to-video (M+V2V) mode, which aligns with our inpainting formulation by applying effect masks to localize edits;  
(c) \textbf{LoRA-Edit}~\cite{gao_lora-edit_2025} — a per-video tuning method that learns localized knowledge from a source video and its corresponding bounding-box mask, then propagates edits from the first edited frame to subsequent frames (I+M+V2V). Since the input composite $\inoeffect$ lacks environmental effects, we apply a full-frame mask to enable knowledge learning across the entire video (I+V2V); and  
(d) \textbf{Runway Aleph\footnote{\href{https://app.runwayml.com/}{https://app.runwayml.com/}}} — a commercial, in-context video-to-video (V2V) editing platform that delivers state-of-the-art visual quality but does not support masked-region editing.  

We present qualitative comparisons in Fig.~\ref{fig:benchmark} for methods that do not support mask guidance, and in Fig.~\ref{fig:control} for those that do. All methods are evaluated using identical text prompts and consistent input conditions. 
Overall, \methodname~produces state-of-the-art realistic effects while faithfully preserving the original video content.

\subsection{Quantitative Comparison}
\label{sec:quantitative_comparison}
We collect 24 test videos for benchmarking, including 18 videos from DAVIS and 6 in-the-wild real-world videos.
Given the ground-truth videos $\igt$ and generated results $\hat{\mathcal{I}}$, we report the average frame-level metrics: CLIP score~\cite{hessel2021clipscore}, SSIM, PSNR, and LPIPS across all frames. In addition, we evaluate video-level metrics including debiased FVD~\cite{ge2024content}, VMAF~\cite{blog_toward_2017}, and VBench (overall consistency)~\cite{huang_vbench_2023_fixed}. For baselines that do not support long videos, we uniformly trim all videos to the same length to ensure a fair comparison.

As shown in Table~\ref{tab:benchmark}, our method achieves state-of-the-art overall performance. However, the generated effects can be subtle or spatially localized, resulting in only marginal gains in conventional CLIP-based metrics. 
A detailed analysis of these limitations is provided in the supplementary material (SM).
To capture such fine-grained perceptual differences, we draw inspiration from recent work on directional evaluation in embedding spaces~\cite{NEURIPS2024_34a9582c_fixed, huggingfaceEvaluatingDiffusion} and propose an appropriate metric, $\text{CLIP}_{\text{dir}}$, which measures \emph{directional cosine similarity} between CLIP embeddings. Formally,
\begin{align}
    \text{CLIP}_{\text{dir}} = 100 \times \frac{(E_{\igt} - E_{\inoeffect}) \cdot (E_{\mathcal{I}} - E_{\inoeffect})}{||E_{\igt} - E_{\inoeffect}||_2 \cdot ||E_{\mathcal{I}} - E_{\inoeffect}||_2},
\end{align}
where $E_{\mathcal{I}}$ denotes the CLIP image embedding of the generated frame, and all embeddings are L2-normalized before computing cosine similarity.
Intuitively, $\text{CLIP}_{\text{dir}}$ quantifies how closely the change from the input without effects $\inoeffect$ to the generated result $\hat{\mathcal{I}}$ aligns with the change from $\inoeffect$ to the ground truth $\igt$. 
Our numerical benchmarking demonstrates \methodname's state-of-the-art effect generation capabilities, both with and without mask guidance $\meffect$.

\subsection{User Study}

\begin{table}[t]
\centering
\caption{
\textbf{Comparison across text, mask, and FG\&BG fidelity.}
`--' denotes baselines without mask guidance, for which we compare only to our unmasked variant (Ours$^\dagger$) for fairness.
User preference (win rate) demonstrates our state-of-the-art qualitative performance across all attributes.
}
\resizebox{\columnwidth}{!}{%
\begin{tabular}{l|cccc}
\toprule
\diagbox[width=10em]{Fidelity to}{Ours vs.}  & AnyV2V & LoRA-Edit & Runway & VACE \\
\midrule
Text   & 92\%  & 70\% & 52\% & 85\%  \\
Mask   & -- & --  & -- & 82\%  \\
FG \& BG  & 98\% & 97\% & 77\% & 79\%  \\
\bottomrule
\end{tabular}%
}
\label{tab:user_study}

\end{table}

\noindent
We conduct a pairwise user study comparing our method against baselines along three axes:  
(a) \textbf{Text fidelity} — ``Which video's effects and details best match the editing prompt?'';  
(b) \textbf{Mask fidelity} — ``Which video's effects and details better align with the masked regions?''; and  
(c) \textbf{Foreground \& background fidelity} — ``Which video best preserves the appearance and motion of the original content while adding new effects?''

Our user study involved 30 participants, including 14 professional VFX artists and 16 non-expert users. 
We collected a total of 1{,}499 responses across four comparison groups: 370 responses over 12 video pairs against AnyV2V~\cite{ku2024anyv2v}, 510 responses over 17 pairs against LoRA-Edit~\cite{gao_lora-edit_2025}, 329 responses over 11 pairs against Runway Aleph, and 290 responses over 10 pairs against VACE~\cite{vace}.
We report the win rate of our method against each baseline in Table~\ref{tab:user_study}.  
Notably, our method achieves comparable overall quality to the state-of-the-art commercial tool (Runway Aleph), but improves adherence to the input video and more explicit control over the output effects.


\subsection{Ablation Study}
\label{sec:ablation}
We evaluate the impact of incorporating the unpaired data introduced in Sec.~\ref{sec:control_prompt}. 
As shown in Fig.~\ref{fig:ablation}, adding this data notably improves the model’s prompt-editing capability. 
Quantitative results in Table~\ref{tab:benchmark} further show that removing the unpaired data yields the lowest $\text{CLIP}_{text}$ score, confirming its importance for preserving language-conditioned generation. 
We further analyze how each dataset source contributes to overall performance in SM.

\begin{figure}[ht]
\vspace{0.75em}
\centering
\begin{overpic}[width=\linewidth]{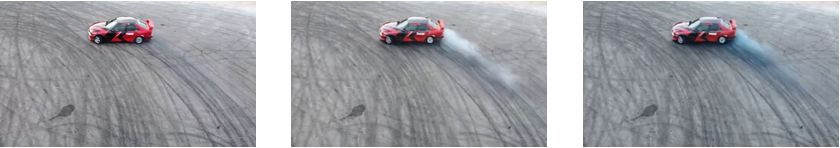}
    \def\ypos{19}
    \put(5, \ypos){\scriptsize Input w/o effects}
    \put(35,\ypos){\scriptsize Output w/o unpaired data}
    \put(70,\ypos){\scriptsize Output w/ unpaired data}
\end{overpic}

\vspace{-0.5em}
\captionof{figure}{
\textbf{Ablation.}
Given an input video of a drifting car without effects, we prompt \methodname~to add \textit{``blue smoke''}.
The ablated model trained with only paired data cannot follow the text prompt, but the full model restores the base model's prompt adherence.
}

\vspace{-1.5em}
\label{fig:ablation}
\end{figure}

\section{Extensions and Applications}
\label{sec:application}
As shown in Fig.~\ref{fig:control}, \methodname~supports versatile workflows, including mask-based control, text-prompt editing, and CFG scaling. 
We further showcase downstream applications of \methodname~across diverse real-world scenarios, highlighting its flexibility and compositing capabilities. 

\noindent
\textbf{Text Prompt Editing.}
As shown in Fig.~\ref{fig:control}, \methodname~supports straightforward text-based edits, such as changing the color of generated smoke. Beyond these modifications, \methodname~also enables fine-grained, subtle effect control with CFG scaling. As shown in Fig.~\ref{fig:application_prompt}, it can generate stylistic variations of the same effect. 

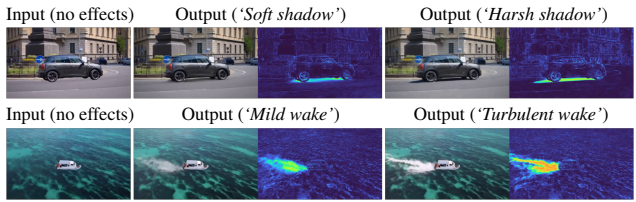
\begin{figure}[ht]
    \vspace{0.75em}
    \centering
    \begin{overpic}[width=\linewidth]{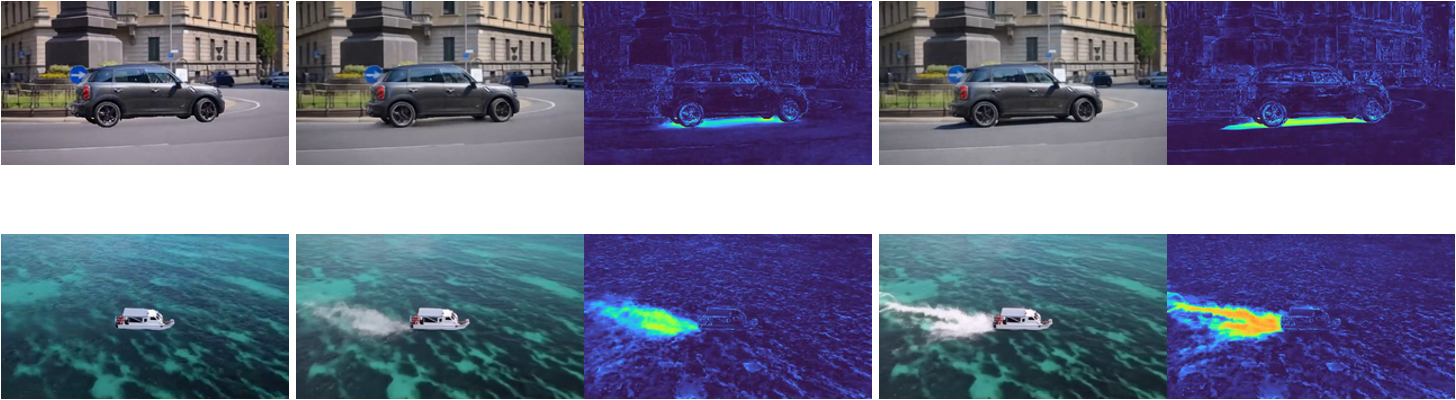}
    
        \def\ypos{28.7}
        \put(0,\ypos){\scriptsize Input (no effects)}
        \put(27,\ypos){\scriptsize Output (\textit{`Soft shadow'})}
        \put(66,\ypos){\scriptsize Output (\textit{`Harsh shadow'})}

        \def\ypos{12.7}
        \put(0,\ypos){\scriptsize Input (no effects)}
        \put(28,\ypos){\scriptsize Output (\textit{`Mild wake'})}
        \put(65,\ypos){\scriptsize Output (\textit{`Turbulent wake'})}
        
    \end{overpic}

    \vspace{-0.5em}
    \caption{
    \textbf{Text prompting for fine-grained control.}
    \methodname{} enables precise modulation of effect intensity and style through textual prompts, with the difference map shown alongside each result. 
    }
    \label{fig:application_prompt}
    \vspace{-0.5em}
\end{figure}

\noindent
\textbf{Robust Mask Editing.}
For non-expert users, manually annotated masks are often imperfect and may include unreasonable regions due to hand-drawing errors. We demonstrate the robustness of \methodname~to such imprecise mask inputs. 
As shown in Fig.~\ref{fig:application_mask}, \methodname~can handle challenging cases, including (a) masks that encompass both the foreground and effect regions, and (b) masks that indicate effects in physically implausible regions. 
These results highlight \methodname's ability to maintain semantic consistency and robustly interpret real-world, imperfect user annotations.

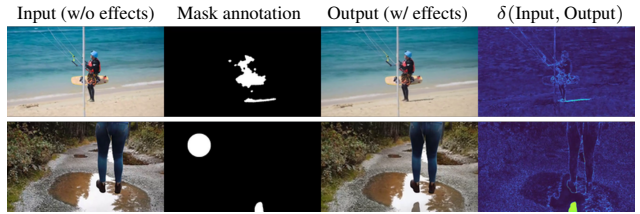
\begin{figure}[ht]
    \centering
    \begin{overpic}[width=\linewidth]{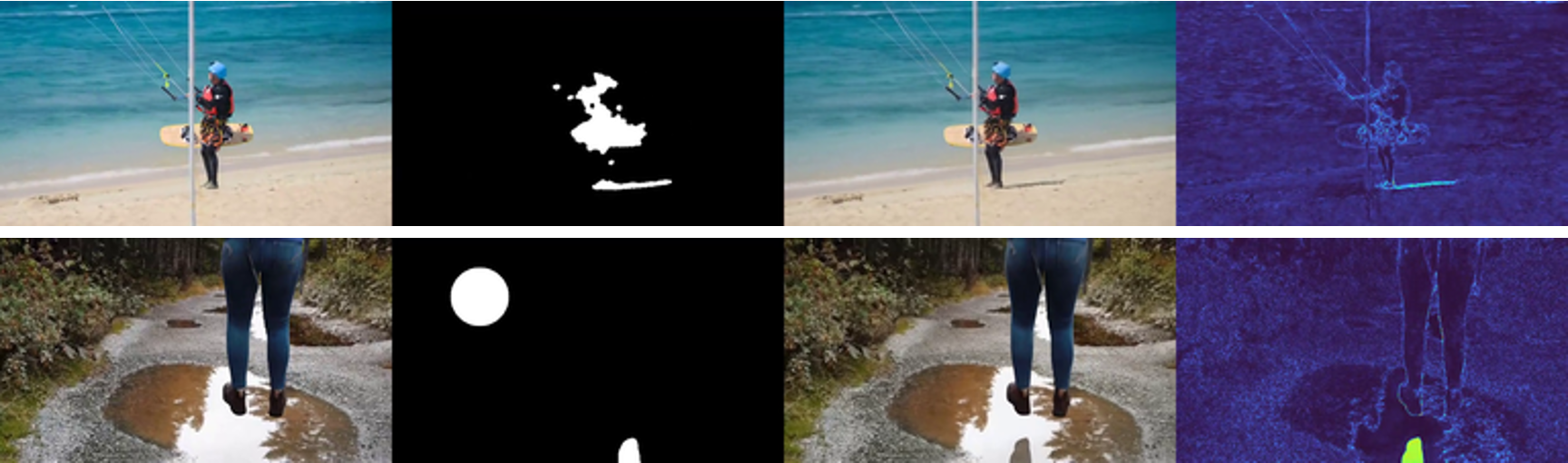}

        \def\ypos{30.7}
        \put(1.5,\ypos){\scriptsize Input (w/o effects)}
        \put(27,\ypos){\scriptsize Mask annotation}
        \put(51,\ypos){\scriptsize Output (w/ effects)}
        \put(78,\ypos){\scriptsize $\delta(\text{Input}, \text{Output})$}
        
    \end{overpic}

    \vspace{-0.5em}
    \caption{
    \textbf{Robustness to imperfect masks.} 
    \textbf{Top:} The input mask includes both the foreground and its shadow, yet \methodname{} only adds a shadow, preserving the foreground. 
    \textbf{Bottom:} \methodname~ ignores the spurious circular mask region and still adds the correct reflection before the person jumps into the puddle.
    }
    \label{fig:application_mask}

    \vspace{-1.5em}
\end{figure}

\noindent
\textbf{Keyframe Mask Annotations.} \methodname~supports keyframe mask annotation, reducing manual user input. As shown in Fig.~\ref{fig:application_keyframe}, \methodname~smoothly interpolates between annotated and unannotated frames, generating shadows and mild water splashes in unannotated frames (w/o keyframe mask), while producing stronger water splashes under the guidance of the annotated keyframes (w/ keyframe mask).



\begin{figure}[ht]
    \vspace{0.75em}
    \centering
    \begin{overpic}[width=\linewidth]{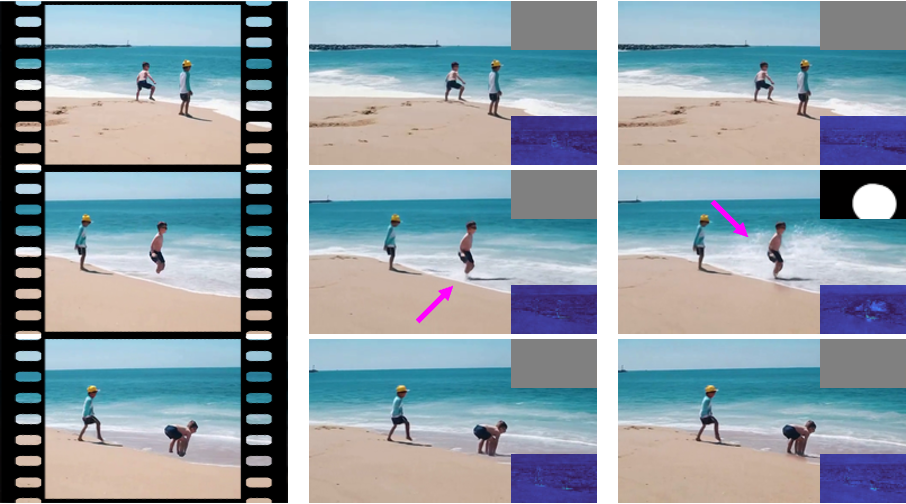}

    
    \def\ypos{57}
    \put(6,\ypos){\scriptsize Input (no effect)}
    \put(38,\ypos){\scriptsize w/o keyframe mask}
    \put(73,\ypos){\scriptsize w/ keyframe mask}

    \def\xpos{-3.5}
    \put(\xpos, 43){\rotatebox{90}{\scriptsize t=10}}
    \put(\xpos, 26){\rotatebox{90}{\scriptsize t=50}}
    \put(\xpos, 7){\rotatebox{90}{\scriptsize t=80}}
    
    \end{overpic}

    \vspace{-0.5em}
    \caption{
    \textbf{Keyframe mask annotations.} 
    \textbf{Left:} Input frames without effects. 
    \textbf{Middle:} Output results without mask guidance (gray). 
    \textbf{Right:} Output results with a single keyframe mask at $t = 50$, while other frames remain unannotated. The keyframe guides the creation of a stronger water splash when the boy jumps into the ocean. 
    Masks are shown at the top-right inset, and difference maps at bottom-right inset.
    }
    \label{fig:application_keyframe}
    \vspace{-1.25em}
\end{figure}
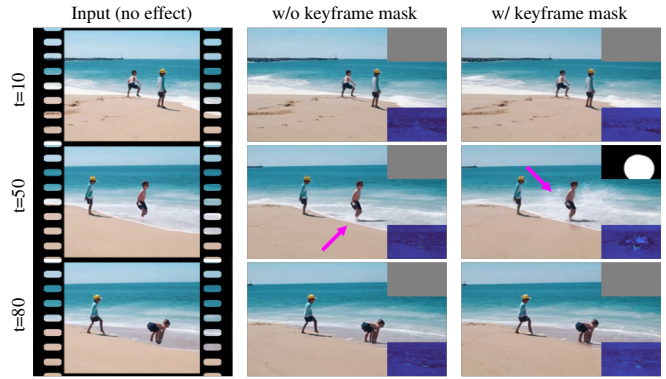



\section{Discussion and Limitations}
While \methodname~achieves state-of-the-art preservation of input content, it may still struggle to produce pixel-perfect reconstructions due to VAE encoding and decoding. Future work could explore per-example test-time optimization~\cite{zhao_objectclear_2025} or incorporate fidelity-enhancement modules during training~\cite{chen_ultrafusion_2025}. 
\methodname~may also hallucinate implausible effects in challenging background regions; fine-tuning on stronger pre-trained priors such as \mbox{Lumiere}~\cite{bar2024lumiere} or Veo3~\cite{veo3} could further improve robustness.

Our method overcomes core limitations of prior generative models for effect generation and, despite not addressing harmonization or relighting, introduces augmented compositing that outperforms previous approaches and supports diverse real-world uses.





\noindent
\textbf{Acknowledgments.}
Thank you to all ILM staff who assisted in preparing this work, especially Miguel Perez Senent for the 3D boat and ocean elements used in Figure~\ref{fig:method} (row 2) and Figure~\ref{fig:control} (row 3), and ILM leaders Rob Bredow, Francois Chardavoine, and Greg Grusby for their assistance in clearing this work for publication.


\vfill \eject 
{
    \small
    \bibliographystyle{ieeenat_fullname}
    \bibliography{main,references_luchao_rebiber}
}

\clearpage
\setcounter{page}{1}
\maketitlesupplementary
\appendix
\renewcommand{\thesection}{\Alph{section}}

\tableofcontents
\vspace{2em}

\noindent
In addition to this supplementary PDF, we provide additional visual materials (e.g., images and videos) on our project webpage at
\href{https://overplusplus.github.io/}{https://overplusplus.github.io/}.
We also encourage readers to refer to the accompanying videos for a more comprehensive evaluation of the visual results.

\vfill \eject
\section{Overview of Appendices}

In this supplemental PDF, we provide the following additional details:
\begin{itemize}
\itemsep0em
\item Sec.~\ref{sup:metrics}: A discussion of the limitations of conventional evaluation metrics, supported by visual examples.
\item Sec.~\ref{sub:ablation}: Ablation studies examining the influence of different training data sources (synthetic versus real).
\item Sec.~\ref{sup:metaprompt_caption}: The system prompt used to generate video captions for our training data.
\item Sec.~\ref{sup:metaprompt_augment}: The system prompt used for caption augmentation to create unpaired text-to-video data, complementing the method described in Sec.~\ref{sec:control_effect}.
\item Sec.~\ref{sup:failure_cases}: Representative failure cases that highlight challenging scenarios.
\end{itemize}

\section{Metrics Analysis}
\label{sup:metrics}

As discussed in Sec.~\ref{sec:quantitative_comparison}, traditional CLIP-based similarity metrics can sometimes be unreliable for evaluating environmental effect edits. In these cases, an image without the inserted effect is scored as more similar to the ground truth in CLIP’s embedding space, while an image with the correct effect (e.g., wake, splash, smoke, or shadow) receives a lower similarity score despite being perceptually more faithful. This phenomenon is shown in Fig.~\ref{fig:sup_metrics}, where CLIP favors visually incomplete outputs simply because they remain closer to the original distribution of the ground-truth image.

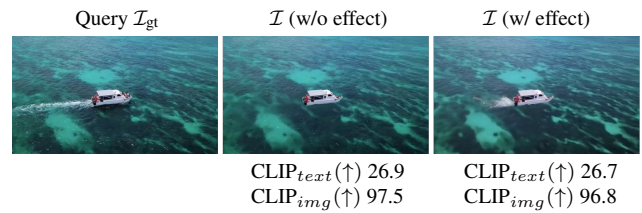
\begin{figure}[ht]
    \vspace{1em}
    \centering
    \begin{overpic}[width=\linewidth]{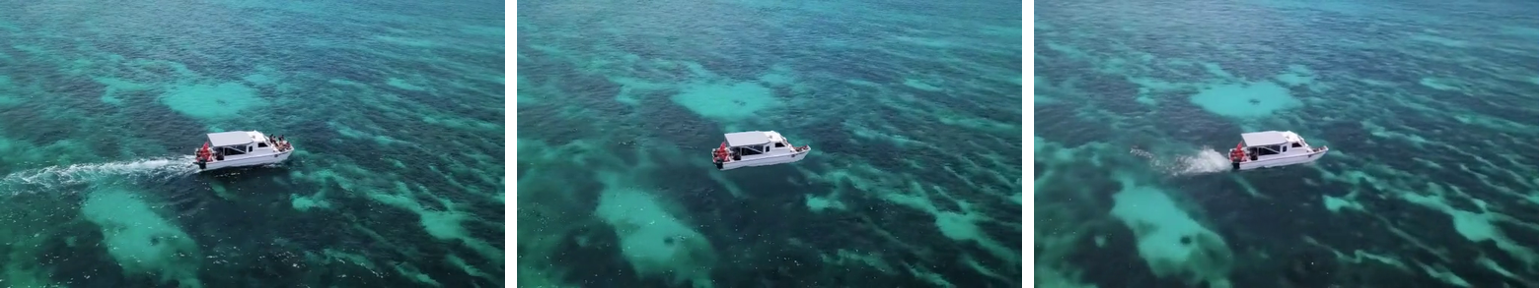}

    \def\ypos{20.5}
    \def\yposi{-4}
    \def\yposii{-8}
    
    \put(10, \ypos){\footnotesize Query $\igt$}

    \def\xpos{38}
    \put(41, \ypos){\footnotesize $\mathcal{I}$ (w/o effect)}
    \put(\xpos, \yposi){\footnotesize $\text{CLIP}_{text} (\uparrow)$ 26.9}
    \put(\xpos, \yposii){\footnotesize $\text{CLIP}_{img}(\uparrow)$ 97.5}

    \def\xpos{72}
    \put(75, \ypos){\footnotesize $\mathcal{I}$ (w/ effect)}
    \put(\xpos, \yposi){\footnotesize $\text{CLIP}_{text} (\uparrow)$ 26.7}
    \put(\xpos, \yposii){\footnotesize $\text{CLIP}_{img}(\uparrow)$ 96.8}
    
    \end{overpic}
    
    \vspace{2em}
    \caption{%
      \textbf{CLIP metric analysis.}  
      Given a reference image $\igt$ and a prompt caption $\mathcal{T}$, we compute CLIP similarity scores for two generated images: one without environmental effects and one with added effects. The results illustrate that CLIP similarity may fail to reflect the introduced environmental effects.%
    }
    \label{fig:sup_metrics}
\end{figure}

\section{Ablations of Training Data}
\label{sub:ablation}

Beyond the unpaired data study in Sec.~\ref{sec:ablation}, we examine the contributions of the other training data sources from Sec.~\ref{sec:training_data} by ablating real and synthetic data. 
Quantitative results are shown in Table~\ref{tab:sup_ablation}, and qualitative comparisons are provided in Fig.~\ref{fig:sup_ablation}.

\begin{figure}[ht]
\noindent
\begin{minipage}{\linewidth}

    \centering

    \captionof{table}{
    \textbf{Ablation study (table).}
    We evaluate the contribution of each data source by removing it from the training set and measuring the drop in performance across three CLIP-based metrics.
    }
    \label{tab:sup_ablation}
    \resizebox{\linewidth}{!}{%
    \begin{tabular}{ccc|ccc}
    \toprule
    \multicolumn{3}{c}{Training data source} & \multicolumn{3}{c}{Metric} \\
    Real & Synthetic & Unpaired & $\text{CLIP}_{dir}$ $\uparrow$ & $\text{CLIP}_{text}$ $\uparrow$ & $\text{CLIP}_{img}$ $\uparrow$\\
    \noalign{\vskip 2pt}\hline\noalign{\vskip 4pt}
    \xmark & \cmark & \cmark & 34.91 & 29.88 & 95.18 \\
    \cmark & \xmark & \cmark & \cellcolor{tabsecond}44.12 & \cellcolor{tabsecond}29.95 & 95.25 \\
    \cmark & \cmark & \xmark & 42.93 & 29.89 & \cellcolor{tabsecond}95.34 \\
    \cmark & \cmark & \cmark & \cellcolor{tabfirst}46.27 & \cellcolor{tabfirst}31.58 & \cellcolor{tabfirst}95.48 \\
    \bottomrule
    \end{tabular}%
    }

    \centering
    
    \vspace{3em}
    \begin{overpic}[width=\linewidth]{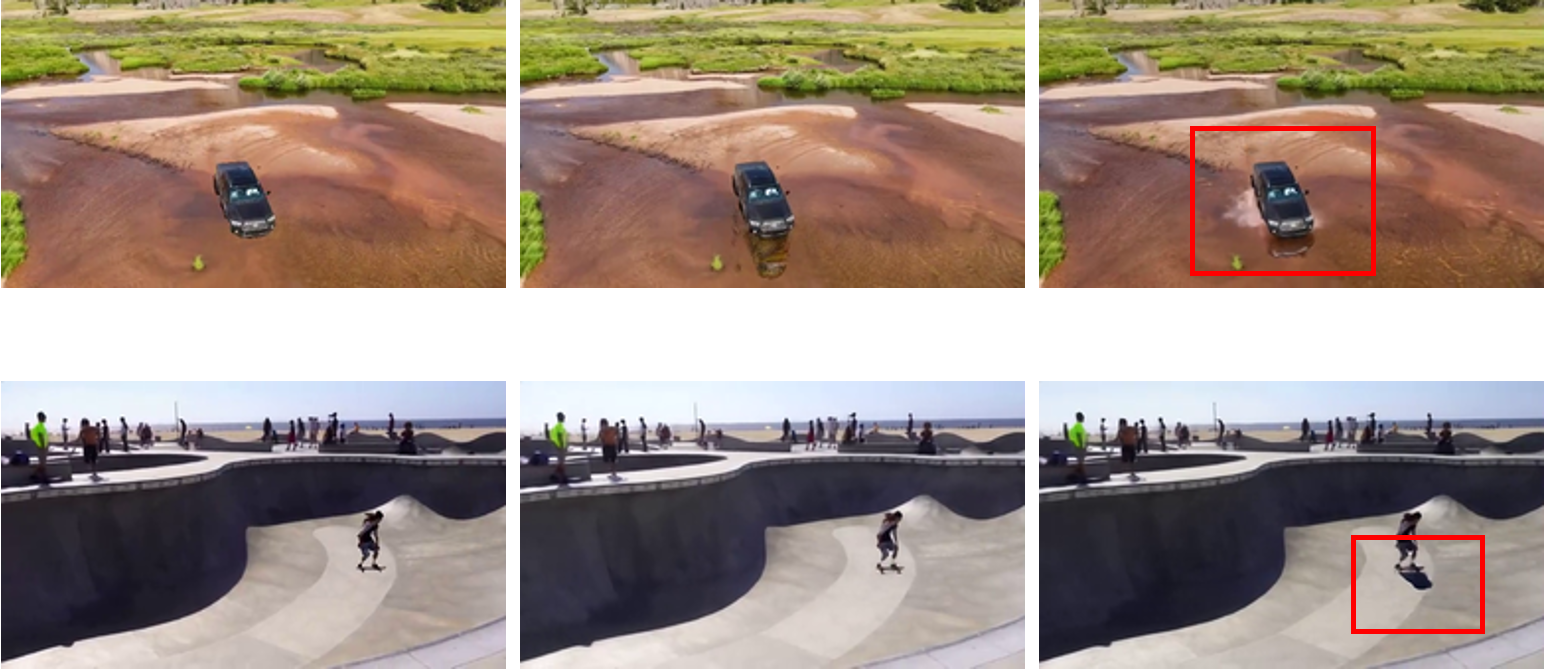}

    \def\yposi{20}
    \def\yposii{45}
    \def\xposi{13}
    \def\xposii{44}
    \def\xposiii{81}

    \put(\xposi, \yposi){\footnotesize $\inoeffect$}
    \put(41, \yposii){\footnotesize w/o Real data}
    \put(\xposiii, \yposi){\footnotesize Full}

    \put(\xposi, \yposii){\footnotesize $\inoeffect$}
    \put(38, \yposi){\footnotesize w/o Synthetic data}
    \put(\xposiii, \yposii){\footnotesize Full}
    
    \end{overpic}

    \captionof{figure}{
    \textbf{Ablation study (figure).}
    Removing synthetic data weakens shadow generation, while removing real data reduces the quality of complex effects such as reflections and water splashes.
    Highlighted outputs correspond to the model trained on the full dataset.
    }
    \label{fig:sup_ablation}

\end{minipage}
\end{figure}

\section{Video Caption Generation}
\label{sup:metaprompt_caption}

We generate video captions $\mathcal{T}$ in a two-stage pipeline. First, we use the VLM MiniCPM-V-2.6~\cite{yao2024minicpm} to produce dense spatio-temporal descriptions that capture fine-grained scene dynamics and environmental interactions. The system prompt used for this stage is provided in Prompt~\ref{sub:prompt_gen1}.

\begin{figure}[th]
    \begin{PromptBox}[label=sub:prompt_gen1]{gray}{System Prompt for Caption Generation}
    Describe the video in detail, with a special focus on environmental effects and how they interact with the scene.
    
    Pay close attention to:
    
    - The **type and intensity** of physical effects — such as soft vs. harsh lighting, subtle vs. violent splashes, thin vs. dense smoke, or gentle vs. strong wind.
    
    - How these effects **interact with the foreground and background** — e.g., shadows cast across surfaces, splashes disrupting water, smoke enveloping objects, or reflections shifting with movement.
    
    - The **spatial extent** (how far the effects spread) and **temporal behavior** (how quickly they appear, move, or dissipate).
    
    Avoid generic summaries. Focus on how **physical forces affect materials** and **how visible cues evolve over time**.
    \end{PromptBox}

    \vspace{-0.6em}
\end{figure}

Next, we refine these dense descriptions with the LLM LLaMA-3.1-8B-Instruct~\cite{grattafiori2024llama}, converting them into concise and coherent video-level captions while preserving the key environmental cues needed for downstream generation. The refinement prompt is shown in Prompt~\ref{sub:prompt_gen2}.

\begin{figure}[ht]

    \begin{PromptBox}[label=sub:prompt_gen2]{gray}{System Prompt for Caption Refinement ($\mathcal{T}$)}
    You are part of a team generating videos with AI models. Your task is to write prompts that guide the model to produce physically plausible, environmentally grounded videos, especially emphasizing **interactions between foreground and background**.
    
    Descriptions should:
    
    - Highlight **environmental effects** (e.g., light, shadow, water, dust, wind, smoke, reflections).
    
    - Emphasize effects **arising from interactions** (e.g., splashes from a foot hitting water, dust clouds from skidding wheels, shadows cast by moving objects on textured surfaces).
    
    - Specify the **degree, intensity, and spatial spread** of effects (e.g., subtle vs. turbulent splashes, soft vs. harsh shadows, thin vs. dense smoke, cloudy vs. glossy reflections).
    
    - Use precise physical language (e.g., "fine mist hangs in the air", "glare flares sharply off wet asphalt", "ripples cascade outward in concentric rings").
    
    - Focus on **visual physics**, **perceptual realism**, and **spatio-temporal interactions**.
    
    Example prompts:
    
    
    - "A race car speeds across a dry desert track, kicking up dense clouds of dust that linger and gradually disperse in golden sunlight."
    

    Limit the prompt to [{x}] words.
    \end{PromptBox}
    
\end{figure}

\section{Video Caption Augmentation}
\label{sup:metaprompt_augment}

As described in Sec.~\ref{sec:control_prompt}, given an initial video caption $\mathcal{T}$ (see Sec.~\ref{sup:metaprompt_caption}), we generate multiple augmented captions to produce diverse unpaired text-to-video (T2V) data.

For example, given the original caption $\mathcal{T} =$ \textit{``A car performs a drift maneuver, unleashing \colorbox{Apricot}{dense, white} smoke in a wide, sweeping arc, gradually dissipating into the air,''}
an augmented caption may be $\mathcal{T} =$ 
\textit{``A car performs a drift maneuver, casting \colorbox{Apricot}{thin, blue-gray} smoke as its tires slide against the asphalt.''}

Below, we provide the system prompt used in GPT-5 for caption augmentation.

\section{Failure Cases}
\label{sup:failure_cases}

We highlight the limitations of \methodname~in Fig.~\ref{fig:sup_failure_cases}. In some cases, the model introduces unintended effects in background regions. We expect that fine-tuning on stronger pre-trained video priors, such as \mbox{Lumiere}~\cite{bar2024lumiere} or Veo3~\cite{veo3}, could further improve robustness.

\begin{figure}[ht]
    \vspace{1em}
    \centering
    \begin{overpic}[width=\linewidth]{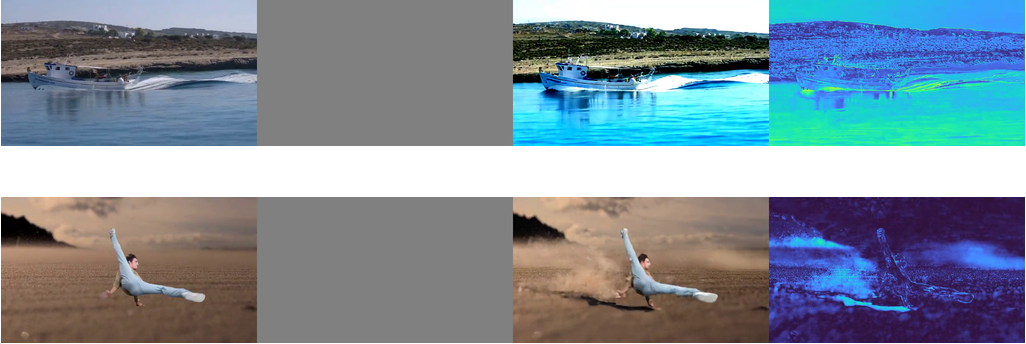}
    
        \def\ypos{35}
        \put(1.5,\ypos){\scriptsize Input (w/o effects)}
        \put(27,\ypos){\scriptsize Mask annotation}
        \put(54,\ypos){\scriptsize Oversaturation}
        \put(78,\ypos){\scriptsize $\delta(\text{Input}, \text{Output})$}

        \def\ypos{15.5 }
        \put(1.5,\ypos){\scriptsize Input (w/o effects)}
        \put(27,\ypos){\scriptsize Mask annotation}
        \put(54,\ypos){\scriptsize Hallucinations}
        \put(78,\ypos){\scriptsize $\delta(\text{Input}, \text{Output})$}
        
    \end{overpic}
    
    \caption{
    \textbf{Failure cases.}
    With excessively high CFG~\cite{sadat_eliminating_2025}, \methodname~may alter the overall color tone of the output  (top).
    The model may also hallucinate unwanted effects  such as dust in challenging background regions (bottom).
    }
    \label{fig:sup_failure_cases}
\end{figure}

\vfill \eject
\begin{figure}[th]
\begin{PromptBox}{gray}{System Prompt for Caption Augmentation ($\mathcal{T}$)}
You are an expert video captioner specializing in visual effects (VFX).

Your task is to generate multiple highly descriptive captions that augment an input caption with **different variations of the same effect type**. Rules:

1. **Identify the effect type** in the input (e.g., splash, smoke, shadow, reflection, fire, dust, ripple, etc.).

2. **Keep the scene, subjects, and actions the same**. Do not replace or remove them.

3. Only change the **effect style**, varying attributes such as:

   - Color (e.g., gray smoke → bright red smoke)
   
   - Intensity (subtle → violent)
   
   - Spatial spread (localized → wide, global)
   
   - Texture / density (thin wispy vs thick heavy)
   
4. Each output caption must be **similar in length** to the input caption.

5. Language should be **natural, vivid, and concise**.

6. Provide **N diverse variations** per input (N is specified by the user, use N=3 if not specified). The variations should be as much diverse as possible. 
\end{PromptBox}

\end{figure}


\end{document}